\documentclass[8pt]{article}
\usepackage{amsmath}
\usepackage{amsfonts}
\usepackage{amssymb}
\usepackage{graphicx}
\usepackage[latin1]{inputenc}
\usepackage{subfigure}
\usepackage{rotating}
\usepackage{dcolumn}% Align table columns on decimal point
\usepackage{bm}% bold math
\usepackage{longtable}
\usepackage{qtree}%Arbres Sintàctics
\usepackage{eepic}%Arbres sintàctics ben definits

\begin{document}
\pagestyle{myheadings}

\markboth{Bernat Corominas-Murtra}{Network statistics on early English Syntax: Structural Criteria}

\title{Network statistics on early English Syntax: Structural criteria}

\author{Bernat Corominas-Murtra$^1$}
\date{$^1$ ICREA-Complex Systems
Lab,  Universitat Pompeu  Fabra,  Dr.  Aiguader  80, 08003  Barcelona,
Spain}

\maketitle

\begin{abstract}
This paper includes a reflection on  the role of networks in the study
of English language acquisition, as  well as a collection of practical
criteria to annotate free-speech corpora from children utterances.  At
the theoretical level, the main  claim of this paper is that syntactic
networks  should be  interpreted  as the  outcome  of the  use of  the
syntactic machinery.   Thus, the intrinsic features  of such machinery
are not accessible directly  from (known) network properties.  Rather,
what  one can see  are the  global patterns  of its  use and,  thus, a
global  view   of  the  power  and  organization   of  the  underlying
grammar. Taking a look into  more practical issues, the paper examines
how to build a net from the projection of syntactic relations.  Recall
that, as  opposed to  adult grammars, early-child  language has  not a
well-defined concept  of structure.   To overcome such  difficulty, we
develop a  set of systematic criteria  assuming constituency hierarchy
and a grammar based on lexico-thematic relations.  At the end, what we
obtain  is a well  defined corpora  annotation that  enables us  i) to
perform  statistics on  the  size of  structures  and ii)  to build  a
network  from  syntactic  relations  over  which we  can  perform  the
standard  measures   of  complexity.   We  also   provide  a  detailed
example.\footnote{\em This paper is  the experimental design of a more
extensive work  {\bf The ontogeny of syntax  networks through Language
Acquisition}, Corominas-Murtra, B., Valverde, S. and Sol\'e, R. V.}.
\begin{footnotesize}
\end{footnotesize}
\end{abstract}

\newpage

\tableofcontents

\newpage

\section{Introduction}

\noindent
In  this  pages  there  is   an  attempt  to  design  and  describe  a
naturalistic experiment  on syntax acquisition.  Specifically, we want
to  build a  {\em Syntactic  network} in  order to  study  syntax with
modern methods of complex network theory.  The process is nor standard
neither straightforward and deserves to be well described.

\noindent
There  are interesting  descriptive  frameworks based  on networks  to
study  syntax.    One  of  them  is  the   so-called  {\em  Dependency
grammar}\cite{Melcuck}. There are,  also, theoretical approaches using
graphs.  A  remarkable member is the  {\em word grammar}\cite{Hudson}.
The approach  assumed here is  closer to the Word-Grammar,  despite we
develop  our  own   criteria,  as  well  as  we   consider  the  graph
representation as a linear projection of the constituency hierarchy.

\noindent
The paper  is organized as follows:  We firstly discuss  the scope and
validity  of the  conceptualization  of syntactic  relations within  a
network.  The  core of the  work is devoted  to the discussion  of the
(descriptive) structural criteria to  tackle the problem of annotation
in early grammars. Finally, a  brief compendium of network measures is
shown, as well as an  illustrating example. All analysis are performed
over the  PETER corpora of  CHILDES database \cite{CHILDES}  using the
DGA-Annotator \cite{DGA}.

\subsection{Different abstractions, different questions: Syntax and Statistical Physics}

\noindent
Every abstraction of a  natural object implies a particular conception
of it  in order  to answer a  specific question.  Assuming  that every
abstraction implies  a simplification, we  have to explore,  then, how
different  approaches can be  complementary or  whether some  of these
approaches  are more  fruitful than  others -i.e.,  what are  the core
questions leading to the  understanding of such phenomena. Focusing on
language, research  on syntax seeks to  find the minimal  set of rules
that could  generate all  -and only- the  potentially infinite  set of
sentences of a given language.  Thus, the question addressed by syntax
is the problem of decidability or computability of the set of possible
sentences  of a  given  language.   When dealing  with  language as  a
complex network, we  have to note that statistical  physics works from
different perspectives:  What are the global features  of the dynamics
of our system? How the combinatorial space is filled? What is -if any-
the role of constraints?

\noindent
Thus,  we don't  address  questions concerning  the  structure of  the
inhabitants  -sentences- of our  system, but  its global  dynamics and
organization. Note that  the questions are different than  in the case
of syntax: thus, the abstraction  we are working in is also different.
Note,  also, that  we  are  not negating  nor  denying the  particular
features of sentence construction.  Simply,  we work at other level of
abstraction.  We  are confident  that information from  this different
level  of approach  should  be enlightening  to  questions addressed  on
grammar itself.

\noindent
If one wants  to apply statistics on some  syntactic phenomena, a word
of  caution is needed  because there  is a  gap between  the syntactic
procedure and the statistical physics procedure: The former is focused
on explaining  almost {\em  every} subtlety of  sentence construction,
while the  latter works on averages  over the largest  possible set of
data. Thus, a compromise has to  be assumed because it is not possible
to deal with  every syntactic phenomena but, also,  the statistics has
to be built on certain criteria.

\subsection{Aims}

\noindent 
Thus, the  aim of  this document  is to present  a set  of descriptive
criteria to identify {\em structure} in early child grammars.  This is
not a theoretical reflection about  the concept {\em structure} or its
evolution  during  the process  of  language  acquisition. With  these
criteria, we  want to build  up the so-called syntactic  networks from
early  child  grammars.  Indeed,  even  though  many  features of  the
language acquisition  process have  been identified and  well studied,
there is a lack of a clear  concept of what it is structured or not in
early  child  grammars,  namely,  there  is  not  a  concept  such  as
grammaticality \cite{Chomsky71} or {\em convergence} \cite{Chomsky95},
defined in  adult grammars.  If  we take the adult-grammar  concept of
grammaticality,  we  will  surely  reject  almost  all  of  children's
productions.   But it will  not be  true that  many of  these rejected
utterances are unstructured at all.

\noindent
In  order  to  overcome  these  limitations, we  developed  a  set  of
descriptive  criteria to  extract the  syntactic network  of different
sets of the child's utterances  belonging to successive time stages of
the language  acquisition process. As  we discussed above,  we present
these  criteria  employed  in   the  construction  of  the  associated
networks\footnote{Note that many properties of the networks make sense
asymptotically, i.e.,  many utterances need  to be analyzed  such that
the results acquire statistically significance.}.

\section{Syntactic networks}

\subsection{From syntax to networks: what we win and what is lost}

\noindent
Formally  speaking, a  given language,  ${\cal L}$  is composed  of an
arbitrary  large, but  finite set  of lexical  items ${\cal  W}$  - or
alphabet, in  technical words-  and of a  restricted set of  rules and
axioms, $\Gamma$.   These rules describe how the  elements from ${\cal
W}$ can  be combined in order  to {\em 1}. obtain  sentences of ${\cal
L}$ or to {\em 2}. decide  if a given sequence of elements from ${\cal
W}$ is  a sentence of  ${\cal L}$ or not  \cite{Chomsky1963b}.  Syntax
properties are, thus, indicators of grammatical complexity\footnote{In
fact, if  we would be able  to design the minimal  program to describe
our system,  its size (in bits)  would be an index  of complexity. See
\cite{Chaitin}}.  If one intends to  develop a syntax theory to decide
whether  a  given sequence  of  words  -namely,  Russian words-  is  a
sentence  of  {\em  Russian}  language,  one needs  to  develop  rules
involving hierarchical and long range relations.  Moreover, the set of
rules must be {\em generative},  in the sense that they should involve
some recursive condition to  grasp the potential infinity of sentences
generated by Russian grammar \cite{Chomsky1963a}.

\noindent
Thus, we  can say,  without any loose  of generalization,  that syntax
works  at  the  {\em  local  level} of  language\footnote{We  are  not
considering  the  usual  {\em  locality}  of  syntactic  relations  as
understood in  many works of  syntax, we use  the term {\em  local} to
specify that syntax operates at  the level of individual elements of a
given language ${\cal L}$ }, i.e.it operates at the sentence level, no
matter  how long  the sentence  is. Now,  we wonder  about  the global
profiles of  syntactic relations.  Note  that the question we  want to
address  is not  finding the  specific  rules needed  to generate  the
possible sentences  of ${\cal L}$, but we  want to take a  look at the
system as  a whole. This could  seem bizarre when  considered from the
point of  view of mainstream  theories of syntax,  but it is  a common
procedure  in  statistical   physics.   Global  profiles  can  provide
information  about general  dynamics and  constraints acting  over the
whole  system as  a complex  entity. The  unexpected profile  given by
Zipf's  law is  an example  of  global behavior  of language  dynamics
\cite{Zipf}.

\noindent
A note  must be  added concerning the  naturalistic character  of this
kind  of   experiments.  Syntax  has  been   related  with  competence
abilities.   But statistical  and naturalistic  works are  carried out
over performance data.  Thus, we  are inferring the global patterns of
performance by assuming some competence abilities.

\subsection{Syntactic Networks}

\noindent
Networks revealed as an  interesting abstraction to explore the global
behavior and dynamics of complex  real systems made from units and the
associated   relations  between  such   units.   Let's   explore  such
abstraction for syntax relations. A network ${\cal G}(V,E)$ is defined
by  the  nodes   $V$  and  the  links  $E$   relating  the  nodes  $V$
\cite{Bollobas}. These  links can be  directed or undirected;  we will
use the directed  ones, if the contrary is not  indicated.  To build a
syntactic  network, the mapping  of ${\cal  L}$ onto  a graph  will be
straightforward for the  set $V\rightarrow {\cal W}$ i.e.,  the set of
lexical  items of  ${\cal  L}$ will  be  the set  of  nodes of  ${\cal
G}$. The  mapping from  $\Gamma$ to  $E$ is not  so obvious  and needs
further considerations.

\subsubsection{From syntactic relations to links}

\noindent
As  we discussed  above, the  syntactic rules  needed to  generate any
natural language revealed considerable degree of complexity.  Thus, it
is  clear   that  the  statistical  treatment  employed   here  is  an
approximation.  Modern syntax is based on recursive operations of {\em
merge}  and {\em  move}  \cite{Chomsky95}.  Such  operations lead  the
syntactic   derivations  to   display  hierarchies   and   long  range
relations. These are features that  cannot be captured explicitly by a
descriptive framework based on linear relations among lexical items -a
network approach.  But we  are approaching the language structure from
the  point of  view of  statistical physics:  we want  to  capture the
global patterns  of the system, thus  we cannot specify  {\em all} the
local properties.  This is contrary  to the procedure employed  in the
Ising models  of ferromagnetism, despite the success  of this approach
is universally acknowledged.  Thus we  have to decide what is the most
essential  structure in a  syntactic derivation.   We assume  that the
most fundamental  thing one can say  from the syntactic  point of view
about a sentence is its {\em constituent structure}.

\noindent
Constituent  structure can be  captured by  linear relations.   In the
following, we define an exact  mapping from a hierarchical binary tree
to a graph\footnote{In the approach of Word-Grammar, the projection of
hierarchical structures into linear dependencies is just the inverse of
what we adopted here. But  it is, essentially, the same procedure. For
more  information,  see  \cite{Hudson}},  an entity  made  of  binary
relations (see figure (\ref{Grafets})):

%\begin{figure}[h] \includegraphics[width=5 cm]{phraseToGraph.eps}
%\caption{From constituent structure to a graph representation of the sentence. Note that {\em the} appears two times in
%the production and derivation. In the graph representation it appears s a single nodes with two links.
%  \label{phraseToGraph}
%}
%\end{figure}

\begin{enumerate}

\item  Find  the basic  syntactic  structure  of constituents  without
labels nor internal operations, with clear distinctions of complements
and the head in every phrase. Detect the verbs in finite forms.

\item Trace an  arc from the complement to the head  of the phrase. If
the complement of a given phrase  is also a phrase, trace and arc from
the head of the internal phrase to the head of the external phrase. We
want to recover the merging order.

\item The head will be the semantically most relevant item.

\item The verbs in finite forms are the head of the sentence. 
\end{enumerate}

\noindent
With the  above criteria, we  make an attempt  to manage data  with the
less aggressive criteria.   Moreover, these assumptions don't constrain
us to  one or other linguistic  school and grasp  reasonably with the
observed syntactic development of children.

\noindent
With this method,  we do not restrict our set of  sentences to the one
generated  by  finite combinatorics.   We  allow  our  sentence to  be
arbitrary long.  Thus,  our model is only finite  because real data is
finite in nature, but it doesn't negate the theoretical possibility of
infinite generativity, a property  expected for any approach of syntax
\cite{ChomskyMiller}.   Moreover,  the  relevance of  the  statistical
physics  properties  generally  is  found in  systems  asymptotically
large.
\begin{center}
\begin{tabular}{|l|}
\hline
\\
\;\;\;CHI:\;\;\; [Telephone go right here]\;\;\;\;\;\;\;\;(...)\;\;\;\;\\
\;\;\;CHI:\;\;\; xxx [need it] [my need it]\;\;\;\;\;\;\;(...)\;\;\;\;\\
\;\;\;CHI:\;\;\; xxx\;\;\;(...)\\
\;\;\;CHI:\;\;\; [Put in there]\\
\\
\\
\Tree [. Telephone  [.VP go [.\; right there ].\;  ].VP  ] +  \Tree [.VP
need it ]+ \Tree [.  My [.VP  need it ].VP  ]+ \Tree[.VP Put  [.\; in there ].\; ]\\
\\
\;\;\;\;\;\;\;\;\;\;\;Right$^{\curvearrowright}$there$^{\curvearrowright}$go$^{\curvearrowleft}$Telephone+
it$^{\curvearrowright}$need$^{\curvearrowleft}$my+in$^{\curvearrowright}$there$^{\curvearrowright}$put\\
\\
\hline
\end{tabular}

\begin{figure}[h]
\includegraphics[width=7.5 cm]{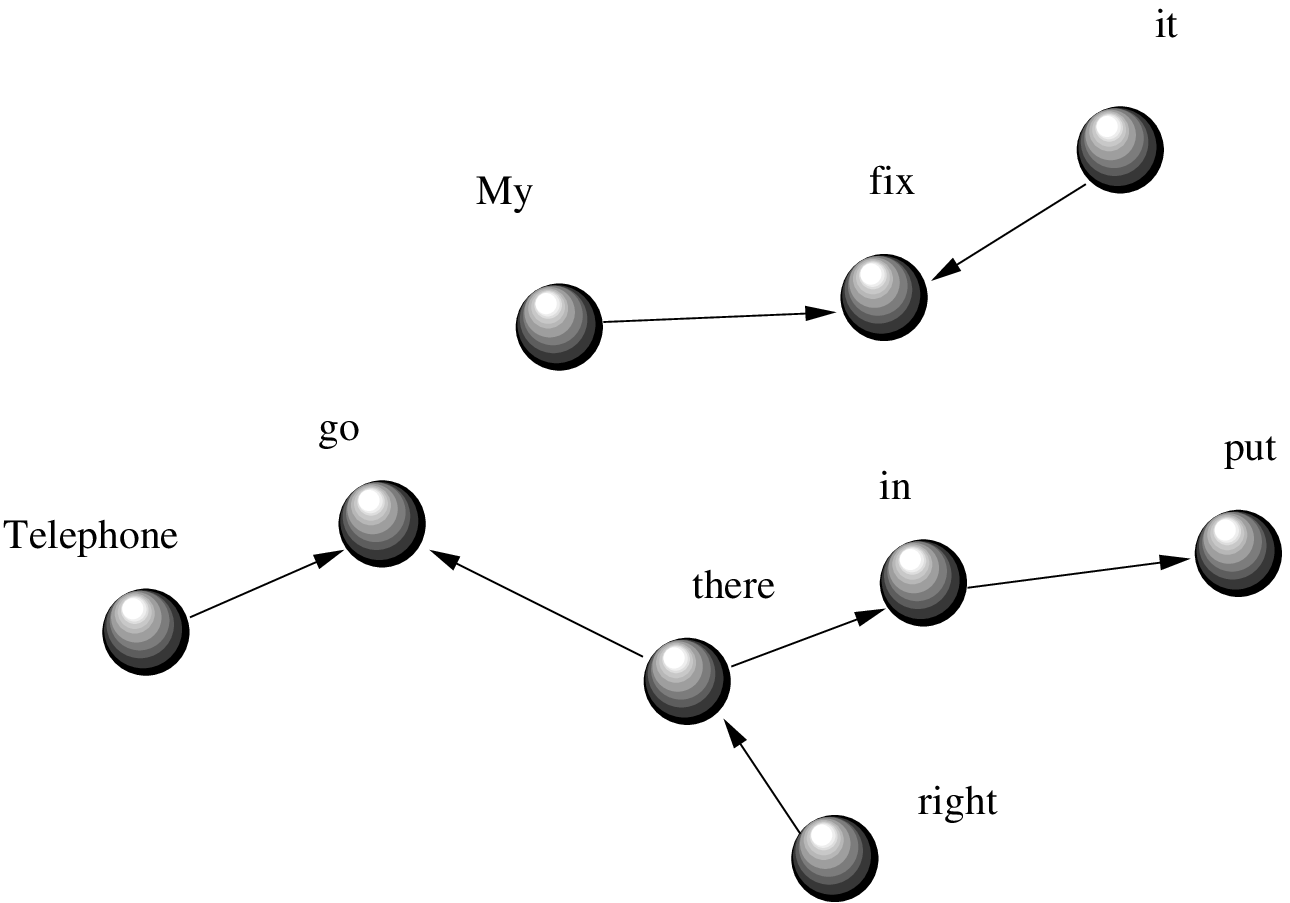}
\caption{\em  Building  syntactic   nets  from  children  free  speech
corpora.  {\bf  A}) We  have the transcript  of a conversation  and we
select only  child's productions.  We identify  the structured strings
. The notion of structure and the used criteria is widely developed in
further  considerations.  {\bf   B})  Basic  analysis  of  constituent
structure, identifying the  verb in finite form (if  any) in different
phrases. {\bf C})Projection of the constituent structures into lexical
dependencies (note  that the operation  is reversible: We  can rebuild
the  tree  from  the  dependency  relations.).  {\bf  D})Following  the
dependency relations found by projecting the naked syntactic structure
we build, finally, the graph.}
\label{Grafets}
\end{figure}
\end{center}

\noindent
Syntactic networks can be built by other procedures. Dependency syntax
\cite{Melcuck} has  been used in other works  \cite{Ferrer04}. In such
an approximation,  syntactic networks have  been built up  by assuming
syntactic   relations   as    dependency   relations   among   lexical
items. Dependency  grammar generates a graph to  describe the sentence
structure  and  it  is the  reason  why  it  is interesting  to  build
networks.     The     mechanism    to    build     large    nets    is
straightforward\footnote{some  authors assume the  network abstraction
for syntax as ontological, i.e.,  not as an approximation to a complex
system  of  rules  involving  recursive  structures  and  non-terminal
nodes(see  \cite{Ferrer04}, \cite{FerrerSyn}).  This  is not  the view
adopted here: The network in our  approach only is an attempt to grasp
some  evolutionary features  of  the system,  properties  that can  be
captured  by taking a  global view  to the  system, something  that is
difficult to achieve when looking  at the local structure of syntactic
relations.  Here, networks  do not substitute the decision/computation
rules  because  some  key  features  of the  syntax  itself,  such  as
constituent hierarchy  or movement, cannot be treated  properly by the
graph theoretic abstraction}.

%\begin{figure}[h]
%\includegraphics[width=8.5 cm]{P5NolinealSimple.ps}
%\caption{A syntactic network of early stages of acquisition. This network belongs to the period when th child grammar
%is going to the functional stage. Some key complexities can be observed, such as the presence of a giant component in the graph where
%almost all the connected words belong, or the emergence of hubs -the most connected words. Links are
%syntactic relations and nodes lexical items. Single nodes are lexical items for which it has not been possible to define any relation
%in the corpus. Data belongs to the corpus {\em Peter5} from Childes database.
%  \label{Peter5Graph}
%}

%\end{figure}

%\begin{figure*}
%\includegraphics[width=13 cm]{EvolChildgrafs.eps}
%\caption{Syntactic graphs made up by using the described criteria. The former and the second belong to the prefunctional stage. Last one, only about a month later, belongs to the functional stage.}
%  \label{Peter6Graph}
%\end{figure*}

\section{Data}

\noindent
Studies on  language acquisition can  be divided into two  main types:
{\em experimental} and {\em  naturalistic}.  The experimental ones are
focused on child's response to well-established situations in order to
obtain data of some specific trait. Naturalistic studies, at the other
hand, are based on child's  free speech corpora.  These corpora can be
extracted, for example, from a recorded session where the child speaks
with  adults  spontaneously.  \cite{Radford}.   Our  study is  clearly
naturalistic, and this label takes here its whole meaning, because the
procedures  to build up  {\em biological}  networks, for  example, are
conceptually, the same.

%Atenci\'o, cal citar apropiadament el Bloom i el Peter.

\noindent
Data   has   been  extracted   from   the   well-known  {\em   CHILDES
Database}\footnote{http://talkbank.org}    \cite{CHILDES,    Childes1,
Childes2}.  The  chosen corpus is  the {\em Peter} Corpus,  from Bloom
1970. We  choose this  data for  many reasons:  1) Time  intervals are
regular  (about 2  or 3  weeks). 2)  Extension of  the corpora  can be
considered  large  enough  to  seize global  properties,  taking  into
account  the intrinsic small  size of  the system.  There is  a little
exception  in  corpus   2,  which  is,  by  far,   the  smallest  one.
Fortunately, this  corpus does not  seem to belong  to a key  stage in
grammar evolution.
%3) Such corpus has been  long studied and many comparative  data related to
%the evolution  of this  specific child is  avaliable {\em  Refs varies
%representatives}.  
3) The  acquisition stages  of  Peter  seem to  be  the standard  ones
observed  in language acquisition.   Thus, it  is reasonable  to think
that  our results  will not  be biased  to strange  deviations  of the
particular case study.

\noindent
Working  data   includes  the  11th   first  corpora  of   Peter's  20
corpora. The age period goes from 1  year and 9 months to 2 year and 4
months.  As we said above, the  aim of the study is to observe whether
and to what  extent syntactic networks can provide  information on the
process of language acquisition.   The so-called {\em syntactic spurt}
(\cite{Radford}), which appears later than the {\em lexical spurt}, is
clearly observable in  the chosen corpora.  Thus, we  manage data that
begins when  the lexical spurt has  already taken place  and ends when
syntactic structures  of child's productions are complex  enough to be
compared with the  adult ones.  This does not  exclude the possibility
of more  abrupt changes  in more advanced  acquisition stages,  but we
stop our analysis here.

\begin{table}
\centering
\begin{tabular}{ccccc}
{\bf Corpus} & {\bf Age} & \phantom{X} & {\bf Corpus} & {\bf Age}\\ 
\hline
1 & 1;9.7 & \phantom{X} & 7 & 2;1.0\\
2 & 1;9.21 & \phantom{X} & 8 & 2;1.21 \\
3 & 1;10.15 & \phantom{X} & 9 & 2;2.14 \\
4 & 1;11.7 & \phantom{X} & 10 & 2;3.0 \\
5 & 1;11.21 & \phantom{X} & 11 & 2;3.21 \\
6 & 2;0.7 & \phantom{X} & \\
\hline
\end{tabular}
\caption{Age of Peter in successive corpora ({\bf years;months.days}).
Data from childes database \cite{CHILDES, Childes1, Childes2}}
\end{table}
$\\$

\noindent
Material contains  several conversations between adults  and the child
(These  adults  are, mainly,  researchers  and  Peter's parents).   We
selected the child's  productions  and  we  studied  them  considering  the
discursive  context where  such utterances  have been  produced.  This
enables us  to {\em clean}  the data.  What  it means is that  we will
discard  1) imitations  from adults  2) non-structured  utterances.  A
complete explanation of the criteria to accept productions and by this
implying that  they contribute to  the syntactic graph is  reported in
the next section {\em Criteria}.

\noindent
A  final   note  concerning  the   data:  it  seems  clear   that  the
morphological nature of English, with poor inflectional features makes
the identification of functional items  easier than in a language with
richer inflectional  features. The global impact  of the morphological
nature  of a  given  language  on network  topology  cannot be  denied
\cite{SoleLPN},  but  the global  reorganization  process observed  in
child    syntactic    networks    seems    to    go    beyond    these
singularities\footnote{obviously,  words as  fundamental  units is  an
intuitive but rather  arbitrary choice. Thus, the same  study could be
extended by  considering morphemes as the fundamental  unit.  This is,
maybe a more  reasonable choice. In this  way, it could
be possible to detect more similarities when comparing the acquisition
processes of different languages.}.

\section{Building the Networks of Syntactic Acquisition: Criteria}

\noindent
We selected the  productions that allow us to  identify some syntactic
structure. Obviously, the  word {\em criteria} is due  to the evidence
that despite  the fact  that most of  early child-productions  are not
grammatical  in the  sense  of full  convergence  or complete  feature
checking, it is  not true that they have no  structure. Thus, the work
of  the  linguist  consists  in  identifying the  clues  of  syntactic
structure in  child's productions.  Selection  is not easy at  all, as
there does not  exist an explicit definition of  syntactic structure in
early grammars.   We considered that  there exists structure  if there
exists, at  least, some lexico-thematic relation  between the elements
in a  production. This  is the basis  of syntactic structure  of early
English  grammars \cite{Radford}.   More complex  relations, involving
functional  words, appear later  and syntactic  structure can  be more
easily identified.  This is coherent with the observed nature of early
grammars.

\subsection{Non accepted productions}

\noindent
First of  all, we  discarded some transcribed  strings if:  1)they are
simply an onomatopoeia with no structural role (in some cases {\em
choo  choo} could  replace {\em  train}). 2)they  are  non transcribed
items -because we supposed it  was not possible to understand what the
child said. We choose not to consider any of these unidentified lexical
items (transcribed in the corpora as  {\em xxx} or {\em yyy}) in order
to ensure the transparency of data managing.
$\\$
\begin{center}
\begin{tabular}{p{8.5cm}}
  \hline
  \vspace{0.1 cm}
   These non-accepted elements are: {\em a} (in some specific contexts), {\em ah, an} (in some specific contexts), {\em awoh, ay, 
    hey, hmm, huh, ka,
    ma, mm, mmhm, oh, oop, oops, ow, s} (in some specific contexts) {\em sh, ssh. ta} (in some specific contexts) {\em uh, uhhuh,
    uhoh, um, whoops, woo, yum}. {\bf Onomatopoeia}: {\em choo, Moo, Woof, Bee Bee}
  \vspace{0.15 cm}
  \\
  \hline
\end{tabular}
\end{center}
$\\$

\noindent
The  case concerning  {\em a},  the schwa,  will receive  a particular
attention below.

\noindent
Some  onomatopoeia  appear   together  with  its  corresponding  lexical
item. To analyze  them, we assume onomatopoeia to  be nonexistent. Take,
for example: $\\$

{\bf Peter 9}{\em I want ta  write the choo choo train $\rightarrow$ I
want ta write the train}

$\\$
\noindent 
Considerations related to other non-trivial interpretations,
such  as  the role  of  {\em ta},  are  extensively  developed in  the
following lines.

\noindent
In addition, and following the enumeration of non-accepted productions,
we  find  the general  case  where no  structure  is  identified in  a
production. In this  situation, we consider the utterance  as a string
of  isolated lexical  items.  Consequently, no  links  but only  nodes
corresponding to the lexical items are added to the graph.

\noindent
More attention has  to be paid to imitations.   The reason to consider
imitations as  unacceptable productions is that we  have no confidence
that such  string is identified as  a structured one or,  simply, as a
single lexical  element.  Imitations  are identified by  analyzing the
discursive context.  Some utterances  of surprising complexity for its
corresponding  stage are  produced after  an {\em  untranscribed adult
conversation}: we cautiously removed from the graph such contributions.
In Peter  5 corpus, {\em  I can't see  it} is produced after  an adult
conversation and  it is, by far,  the most complex  production of this
corpus. It strongly suggests that  this is an imitation from something
said in such untranscribed adult conversation.

\subsection{Accepted Productions}

\noindent
As we stated above, structured productions and lexical items are taken
into  account.  Now we  state  another  assumption:  If in  the  whole
utterance  we  cannot  find   global  structure  but  there  are  some
structured strings, then we take these structured strings separately,
(see figure \ref{Grafets}).

%\begin{center}
%\begin{figure}[h]
%\includegraphics[width=3 cm]{substructure.eps}
%\caption{An utterance partially structured: object appears two times: pronominalized, {\em it}; and explicitly, {\em car}. We
%consider thematic relations between {\em it} and the verb {\em screw}. {\em Car} is accepted as a single lexical item.
%  \label{substructure}
%}
%\end{figure}
%\end{center}

\subsubsection{ Phrases and missing arguments}

\noindent
In the pre-functional stage (identified in our data until corpus Peter
6) there appear a lot of utterances where only thematic relations seem
to  be considered  by the  syntactic  system of  the child.   Thematic
relations are fundamental at the syntactic level, and their appearance
indicates   presence   of   sub-categorization  mechanisms   in   child
grammar. No traces of more complex structure -like agreement- is found
in this early stage of acquisition. We consider as syntactic relations
the thematic relations between  verb and arguments.  Moreover, subject
elision is  usual, due mainly to  the facts that  1)utterances are in
imperative mode or 2)there is  no fixation yet of parametric variation
associated to the explicit presence of subject in English.
%canviar
Productions of  this kind are:
$\\$

{\bf Peter 5} {\em  Open box}, instead  of {\em  Open the  box}, (the  determiner is
missing.)
$\\$

{\bf Peter 5} {\em wheel walk} instead of {\em The wheel walks}, (3-singular English
agreement is missing)
$\\$

{\bf Peter 6} {\em two truck} instead of {\em two trucks}, (no  plural   agreement)   

$\\$

%(see   figure \ref{missingCat})

%\begin{figure}[h]
%  \includegraphics[width=8.5 cm]{missingCat.eps}
%  \caption{In the generativist maturational hypotesis, we should think that: in (a) we have a tree of child production where only 
%    lexical-thematic relations are taken into account. When child competence
%    reaches the adult one, the functional categories {\em T} and {\em SD} make possible a more complex syntactic structure, involving
%    functional lexical items, {\em the}, agreement and case assignation.
%    \label{substructure}
% }
%\end{figure}

\noindent
This leads to the logical  conclusion that productions like {\em *open
the} will not be accepted.  The reason is clear: if we assume thematic
relations  as   the  basic  building  blocks  of   child  syntax,  the
non-presence of the semantically  required argument but its determiner
is not enough to define any relation.

\noindent
Relations between  verbal head  and functional words  are specifically
considered in phrasal verbs.  Its isolated production is considered a
structured  utterance. Several  reasons support  our choice:  1) Their
intrinsic  complex  nature,  2)  We  cannot  conclude  that  there  are
lexicalized  imitations  because,  in  adult  speech,  phrasal  verbs
usually are {\em broken} by a noun or determiner phrase: $\\$

{\em Turn [the wheel]$_{SD}$ out}.
$\\$

\subsubsection{{\em To be} verb}

\noindent
Semantically vacuous predications (those which involve the {\em to be}
verb) are often produced without realization the verb.  We argued that
missing arguments  or lack of agreement  in a production  could be not
the only reasons to conclude that  there is not any structure in child
utterances. This was justified  because strong semantically items were
present   in   discussed   productions.    The  case   of   copulative
constructions  will be  treated close  to the  ones  involving missing
functional  words. In this  case, no  presence of  the verb  does not
motivate   the   consideration   of  non-structured   production.   An
interesting production is: 
$\\$

{\bf Peter 5} {\em Wheels mine} Instead of {\em {\bf The} wheels {\bf are} mine}

$\\$
\noindent
In  this case we  have a  predication {\em  mine} from  something {\em
Wheels}.  Formally,  {\em are} is a semantic  link between predication
and the  element from which  something is predicated  \cite{Heinz}. So
the missing of  the {\em to be} verb could  be considered analogous to
the missing of a functional  particle. The same situation arises from:
$\\$

{\bf Peter 7} {\em That my pen}

$\\$
Usually, when inflectional morphology appears, some infinite forms are
present without the finite form of  the {\em to be} verb.  This is the
case of some present continuous utterances such as:
$\\$

{\bf Peter 8} {\em I writting}

$\\$
This  case  should  be  treated  as  the  above  case:  There  is  some
predication with semantic structure.  Just the opposite is also found:
presence of the {\em to be} verb with an infinitive or finite form:
$\\$

{\bf Peter 8} {\em I'm write too}

$\\$
In this case, we could  assume that the child is acquiring inflectional
morphology and that this utterance is a present continuous one without
inflection.  In  the other hand, we  could consider that  {\em 'm} has
not a role in  the sentence and thus, this can be  treated as a single
finite sentence {\em I write too}.

\noindent
Analogously,
$\\$

{\bf Peter 7} {\em I'm do it}

\noindent
or

{\bf Peter 6} {\em cars goes away}

$\\$ Are treated  as single finite sentences: {\em I  do it} and {\em
The cars go away}

\noindent
Some lexicalized phrases in adult language, such as {\em back seat} or
{\em thank  you}, have been  considered as a complex  structures.  The
reason is to be coherent: If we assume that {\em fix it} is clearly an
imperative structured sentence, at this stages of acquisition there is
no reason to think that {\em back  seat} or {\em thank you} have to be
considered differently. Moreover, this interpretation is also coherent
with the one developed for phrasal verbs.

\noindent
A special  case of imitations involving  the {\em to be}  verb will be
accepted.  These imitations  involve  some {\em  adaptation} of  adult
syntax  to the  syntax in  which the  child is  competent.  An example
should be:
$\\$

{\bf (Peter 6)}

{\bf Adult}: {\em Is that a truck?}
$\\$

{\bf Child}$(1)$: {\em That's a truck?}
$\\$

{\bf Child}$(2)$: {\em That a truck?}
$\\$

\noindent
In this  example, adult  production involve an  interrogative sentence
with  subject  inversion.   The  first  imitation  ({\em  Child}$(1)$)
retains  all the  lexical  items  but the  sentence  is translated  as
interrogative  without subject  inversion.  In  the  second successive
imitation ({\em Child}$(2)$) the verb  is missing. But the elements to
define  a predication  are still  at  work -with  a {\em  schwa} as  a
determiner, suggesting  that the child  is entering into  the functional
stage.

\subsubsection{Infra-specification and semantic extension of lexical items}

\noindent
During  the acquisition process,  extension of  meaning is  subject to
variations.  To know which is the intrinsic nature of these changes is
not our  aim, but  we have  to manage such  situations. Thus,  we find
utterances where  the child uses in  the {\em wrong}  way some lexical
item that could  be related semantically with the  {\em right} lexical
item. As an example: $\\$

{\bf Peter 5} {\em More screwdriver}
 
$\\$  
Which  could,  checking  the  context, be  properly  replaced  by
constituents or lexical items with related meanings: $\\$

{\em Another screwdriver}

$\\$
or

{\em screw it again}

$\\$
or

{\em screw it more} (or {\em harder}...)

$\\$
In the first case, we could  consider that the child made some semantic
extension of the word {\em more}  and it has enough traces to define a
syntactic  relation. But  context can  lead us  to a  second  or third
interpretation.  Generally,  if there is  a great ambiguity  we reject
such  utterances as  structured  ones.  In this  case,  we should  not
consider any  syntactic structure. Thus, we don't  define any relation
in productions such as:
$\\$

{\bf Peter 5} {\em Screwdriver help}
$\\$

{\bf Peter 5} {\em More [fix it]$_{SV}$} (We don't define any relation between

$\;\;\;\;\;\;${\em More} and the SD {\em fix it})

$\\$ The semantics of the  productions are intuitive, but is hard to
justify clearly some kind of syntactic dependency.

\noindent
Strings  displaying  mistakes in  the  use  of  personal pronouns  and
possessives  have  considered  as  structured.   Generally,  we  could
associate such mistakes to the absence or weakness of case system. But
many  productions, as we  reported above,  have structure  without any
trace of  case assignation. Examples  of this kind of  utterances are:
$\\$

{\bf Peter 5} {\em My fix it}$\;\;\;\;\;\;\;\;\;\;$instead of {\em I fix it} 
$\\$

{\bf Peter 8} {\em Me write}$\;\;\;\;\;$ instead of {\em I write}
$\\$

\noindent  
This assumption is  reinforced by  realizing that,  in some
cases, a  production with wrong pronoun is  repeated correctly without
any conversational pause:
$\\$

{\bf Peter 8} {\em Me found it (...) I find it}
$\\$

\noindent 
This situation cannot be confused with  the missing of the {\em to be}
verb  such in  the case  of {\em  wheels mine}.  This case  has  to be
considered as  above mentioned when  dealing with missing {\em  to be}
verb structures.

\subsubsection{First functional particles}

\noindent
In   early  corpora   (1-4)  child   productions  display   very  poor
structures. This is the so-called {\em pre-functional} stage, where no
functional words appear in  structured productions. Beyond this point,
some lexical elements  -we are mainly talking about  the $a$, the {\em
schwa}- seem to act as a {\em proto}functional particles. Whether this
schwa has a phonological  or functional-syntactic character is an open
question \cite{Veneziano, Bloom70}.

\noindent
Some  authors related  the  presence of  these  items as  one step  to
combinatorial speech \cite{Bloom70}, but  they realized that, in early
stages,  the role  of  these  items is  more  related to  phonological
processes   of  language  acquisition,   without  any   functional  or
structural  role,  at the  syntactic  level.   Other  authors such  as
Veneziano   $\&$   Sinclair  ``{\em   linked   these  phenomena   more
specifically  to  the  child's  development of  grammatical  morphemes
considering  them as  a sort  of an  intermediate form  on the  way to
grammatical morphemes.}''   \cite{Veneziano}pp 463.  Roughly speaking,
we can say that the core of  this reasoning is rooted in the idea that
the role of such items is  dynamic, going at very first stages as {\em
filler syllabes} without any  syntactic role and acquiring grammatical
features  during the  process  to end  as  functional particles,  with
specific syntactic role.

\noindent
The  lack of  consensus around  a topic  that seems  to be  crucial in
syntactic acquisition theorizations forces us to be really cautious in
interpreting such  items.  Furthermore, functional words  such $a$ are
strongly  candidates to  be the  hubs in  a fully  developed syntactic
network. Hub  are the most connected  nodes on a  network, being, thus
core pieces in network organization.  Every candidate belonging to the
set of functional  particles is specially analyzed in  order to discard
simple  phonological phenomena.   Thus, for  every occurrence  of such
items  there will  be an  individual decision,  taking in  account the
context  and with  the framework  defined by  Veneziano  $\&$ Sinclair
\footnote{In the Veneziano $\&$  Sinclair's study, the chosen language
is  French, but  we  take as  general  some conclusions  that seem  to
coincide with the observed phenomena in English acquisition}.

\noindent
Specifically,  we considered that  sometimes the  {\em schwa}  plays a
functional role. It is reasonable  to assume, thus, that sometimes the
{\em schwa}  is substituting a specific functional  particle.  In this
cases  we assume  that  the  {\em schwa}  acts  within the  syntactic
structure   as  the  substituted   particle.   Several   examples  can
illustrate such reasoning: $\\$

{\bf Peter 6} {\em Light {\bf a} hall}
$\\$

{\bf Peter 6} {\em light in {\bf a} hall}
$\\$

{\bf Peter 6} {\em look {\bf a} people}
$\\$

\noindent  In  this case,  it  seems  that  {\em a}  substitutes  {\em
the}. {\em {\bf a}} should be  treated as a determiner. This is a very
difficult choice, because  purely phonological interpretation could be
enough to justify the presence, specially in the third case.

\noindent
Sometimes choice is really ambiguous. Take for example:
$\\$

{\bf Peter 6} {\em There {\bf a} new one}
$\\$

\noindent Such a case {\em  {\bf a}} could  be easily interpreted  as a
pure phonological  phenomena. But if we consider  the vacuous semantic
nature of the {\em to  be} verb, we could understand these occurrences
as protofunctionals. We removed these most ambiguous cases.

\noindent
We  also  rejected as  unstructured  utterances productions  involving
confuse sequences of functional particles as:
$\\$

{\bf Peter 6} {\em Will an a in there}
$\\$

\noindent
Any interpretation is really confusing.

\noindent
There are  cases where  the presence  of the {\em  {\bf a}}  is clearly
purely phonological.  For example:
$\\$

{\bf Peter 6} {\em more get {\bf a} more}
$\\$

{\bf Peter 6} {\em {\bf a} ride a horsie}
$\\$

{\bf Peter 5}{\em {\bf a} this thumb}
$\\$

{\bf Peter 7} {\em hmmm my a}
$\\$

\noindent Beyond  the non-definition of  personal pronouns due  to the
weakness of  the case system, we find  the pronoun {\em I}  as an {\em
a}: $\\$

{\bf Peter 7} {\em a want milk}
$\\$

{\bf Peter 7} {\em a want ta get out}
$\\$

\noindent
An interesting sequence of that reinforces our considerations is: 
$\\$

{\bf Peter 7} {\em a put it on (...) my put it on}
$\\$

\noindent
Finally, it is interesting to  note the presence of elements that are,
to some extent,  a mixing between {\em a} and {\em  to}: {\em ta}. The
occurrence of this particle is rare and located explicitly at the very
beginning of  the funcional  stage.  The remarkable  fact lies  on the
evidence  that is  located where  it  should be  the preposition  {\em
to}. This could  imply that in fact there is a  transition from a pure
phonological role to a functional one.  $\\$

{\em a $\rightarrow$ ta $\rightarrow$ to} 
$\\$

\noindent 
Thus, we interpret  {\em ta} as an intermediate stage  but, due to its
location within the sentence and  the context, we assume it behaves as
a preposition: 
$\\$

{\bf Peter 6} {\em [Have [ta [screw it]]$_{PP}$]}
$\\$

{\bf Peter 7} {\em [Have[ta [screw it]]$_{PP}$]}
$\\$

\noindent  
The emergence of  English syntax is strongly tied  to the emergence of
functional particles.  This is the  reason why we decided to take into
account  this  kind  of   lexical  components:  despite  almost  every
utterance involving  such items can be object  of many considerations,
there are enough motivation to try to define a descriptive criteria to
deal with them.

\subsubsection{Duplication of functional words}

\noindent
It is usual  to find, at the beginning of the  functional stage, that a
verb that sub-categorizes, for example, a prepositional phrase, display
two successive prepositions: 
$\\$

{\bf Peter 6} {\em Look {\bf at in} there} 
$\\$

\noindent 
To  manage this  kind of  productions  we assumed,  first, that  these
imply that the  child conceives\footnote{{\em Conceives} implies that
the child  is competent in  this ind of  productions, thus we  are not
using this verb in terms of explicit knowledge } a syntactic structure
that  involves prepositional  phrases.  Following  this  reasoning, we
make the following structural description: 
$\\$

{\em Look {\bf at in}  there} $\rightarrow$
{\em [Look [at there]$_{PP}$]}
$\\$

\noindent  In that  case, {\em  in} is  interpreted as  an independent
lexical item.  Not only prepositions are involved  in this duplication
phenomena, but also determiners:
$\\$

{\bf Peter 9} {\em One that screwdriver}
$\\$

\noindent  Interpretation rules  out  {\em  one} as  a  member of  any
structure, leading the SD {\em [that screwdriver]$_{SD}$} alone.  

\noindent
A situation analogous to famous one described by Braine (p.160-161) \cite{Braine} is:
$\\$

{\bf Peter 7} {\em Get {\bf another} one paper $\rightarrow$ Get another paper}
$\\$

\noindent Thus, as above,  determiner duplication is not considered in
the structural analysis.

\subsubsection{Non-structural lexical items}

\noindent
By this  name, we designate  the lexical items  that are present  in a
conversational  framework  but  cannot  be explicitly  interpreted  as
members  of some  syntactic  structure, such  as  {\em Hello},  or{\em
Ok}. The reason to include  these elements as connected to the network
is   due  mainly   because   their  are   produced  in   non-arbitrary
context. Thus, we assume that they linked to the first element of the
sentence they precede:
$\\$

{\bf Peter 5} {\em Ok Patsy}
$\\$  

\noindent Obviously,  previous reasons are  at work when  dealing with
such items.  Thus, the  conversational context has  to be  analyzed to
interpret these  items. For  example, In the  following situation,
{\em bye}  has not been considered  as a member of  any structure. The
reason is that it is  produced among analogous expressions, leading it
to interpret more in a pragmatic sense:
$\\$

{\bf Peter 7} {\em see you, bye, see you}
$\\$

\noindent Sequences  of numbers or  other elements produced as  a list
are not considered as members of any structure:
$\\$

{\bf Peter 7} {\em one two three...}
$\\$

\noindent  Sometimes, personal  nouns  are produced  by  the child  to
demand attention  from adult  people. In these  situations, we  do not
accept them as a members of structured sentences.

\noindent
Some residual cases to be  commented are the ones related with strings
of nouns:
$\\$

{\bf Peter 9} {\em Piece tape...}
$\\$

\noindent  Which are clearly  unstructured, if  conversational context
does not conspire in the other way.  Sequences like
$\\$

{\bf Peter 9} {\em off on tv...}
$\\$

\noindent  are ruled  out  as structured  ones  because any  structure
proposal leads us to a  certainly bizarre sentence in terms of meaning
and because there is a lack of  many elements that can act as clues to
find  some structure. Thus,  they are  considered as  isolated lexical
items: $\\$

{\bf Peter 7} {\em {\bf An} Jenny}
$\\$

\subsubsection{Negation Structures}

\noindent  When the functional  stage is  being consolidated,  we find
more  complex  structures.   Among  others,  interrogatives  involving
subject   inversion  or   negation  structures.

\noindent  Negation structures  sometimes imply  the presence  of the
auxiliary  {\em to  do}  are produced  using  the negative  particle
alone:
$\\$

{\bf Peter 7} {\bf No} {\em put it here}
$\\$

\noindent This  context suggests us  that {\em No} could  be replacing
the  auxiliary  form {\em  don't}.  {\em Don't  put  it  here} is  its
grammatical  counterpart.   Nevertheless,  we  consider  {\em  no}  as
replacing {\em don't} and, thus, as a member of a bigger syntactically
structures  utterance.  Obviously, as  we said  above, context  has to
rule  out  interpretations  such  as  {\em {\bf  No},  put  it  here}.
Analogous structures can be: $\\$

{\bf Peter 7} {\em {\bf No} ride a bike}
$\\$

\noindent There are other  situations where we considered suitable not
to consider {\em no} as a member of any structure:
$\\$

{\bf Peter 8} {\em in the bag {\bf no}}
$\\$

\noindent We cannot conclude that  there is a syntactic relation among
the negation operator and some other lexical item of the string. Maybe
a  parametric \cite{Hyams}  hypotheses  could save  this production  by
suggesting  that  the location  of  the  negation  operator within  the
structure may be a  parametric feature. Despite interesting, we choose
the rule these productions  out for reliability purposes. Furthermore,
at the same time, there are productions like:
$\\$

{\bf Peter 9} {\em {\bf No} in this box}
$\\$

\noindent  suggesting  that  the  child  {\em knows}  the  ordering  of
negation  structures in  English. In  the latter  case, as  before, we
considered not  risky to  identify {\em  no} as a  member of  a bigger
structures sentence.

\section{Corpora Annotation}

\noindent
Previous set of criteria enables us:

\begin{enumerate}

\item
To identify and characterize syntactic structures in early child language.

\item
To project them  into word-word dependencies in order  to annotate the
corpus.
\end{enumerate}

\noindent
The corpora is annotated {\em by hand}. This enables us to be accurate
and to  manage ambiguous  situations. The program  used to  perform the
annotation is  the so-called  {\bf Dependency Grammar  Annotator} (DGA
annotator).  This  program was developed by  Marius Popescu \cite{DGA}
from the University of Bucaresti and has a nice and easy interface. It
works with  XML files, whose  internal structure will be  described in
the example of the last section.

\section{The average size of structures, $\langle{\cal S}\rangle$}

\noindent
With these criteria in hand, we  are ready to perform a first analysis
of grammar  complexity. Such  an analysis is  closed to  the classical
MLU\footnote{Medium lenght of  utterances, often measured on utterance
size in words  or morphemes}. What we can compute,  now is the average
size  of syntactic  structures. Thus,  in  a production,  we can,  for
example,   find  two   syntactically  unrelated   structures$s_1$  and
$s_2s$. The  number of lexical items  of these structures  will be its
sizes $|s_1|$  and  $|s_2|$.  Such  an  utterance  will
contribute  to the  computation of  $\langle{\cal S}\rangle$  with two
structures.

\noindent
For example, in
$\\$

{\em Look {\bf at in} that} 
$\\$

\noindent
we have two structures:

$s_1=$[Look, [at, that]] $\rightarrow \;|s_1|=3$ 

$s_2=$in $\rightarrow \;|s_2|=1$ 

$\langle s \rangle =(3+1)/2=2$

\noindent  (Note   that  single  words  are   considered  as  size-$1$
structures) To obtain $\langle{\cal S}\rangle$ The average is computed
over all  utterances. Such a measure  will provide us  clues to decide
whether  the  size of  productions  has  information about  grammatical
complexity.  Its   evolution  can  be  related   with  working  memory
limitations.

\section{Building the Network}

\noindent
Once we analyzed a conversation, we can build the network. The process
is  as follows. Due  to the  nature of  our analysis,  we will  have a
collection of words, which define the set ${\cal W}$:

\begin{equation}
{\cal W}=\{car, it, ...\}=\{w_1, w_2,..w_n\}
\end{equation}

\noindent This define the set of nodes of our network.  If, during the
conversation, we  find some structure  where two words $w_i,  w_k$ are
related  syntactically -using the  above criteria!-  we say  that $w_i
rightarrow   w_k$\footnote{Do  not   confuse  it   with   the  logical
conditional} and that there is a link $w_i \rightarrow w_k$.

\begin{eqnarray}
{\cal E}=\{car\rightarrow want, it\rightarrow want, it \rightarrow fix...\}= \nonumber \\
\{w_1\rightarrow w_k, w_2\rightarrow w_k,...\} 
\end{eqnarray}

\noindent
Remark that:

\noindent  All the  words and  links only  appear once  a  time.  This
enables us to separate -as far as possible- some contextual deviations
from  the specific  conversations.  Also, there  can  be many  isolated
nodes.

\noindent Finally,  we compute  the adjacency matrix  ${\cal A}_{ij}$.
This matrix is the representation of the graph and the abstract object
where all computations of graph complexity are performed. If the child
produced $n$ different words during the conversation, the size of this
matrix will be, obviously $n^2$.

\noindent
The adjacency matrix of the directed graph will be:

\begin{equation}
{\cal A}_{ij}=\left\{
\begin{array}{ll}
1 \leftrightarrow w_i\rightarrow w_i \\
0\; {\rm otherwise}
\end{array}
\right.
\end{equation}

\noindent
If we consider the undirected version of this graph, ${\cal A}^{u}_{ij}$ will be defined as:

\begin{equation}
{\cal A}^u_{ij}=\left\{
\begin{array}{ll}
1 \leftrightarrow w_i\rightarrow w_j or w_j\rightarrow w_i\\
0\; {\rm otherwise}
\end{array}
\right.
\end{equation}

\noindent
Note that ${\cal A}^{u}_{ij}$ is symmetrical, whereas ${\cal A}_{ij}$ it is not.

\noindent  Now we  are  ready  to perform  an  exhaustive analysis  of
network complexity.

%\appendix
\subsection{Measures}

\noindent
A  first and  fundamental  question  we find  when  dealing with  such
measures  is whether  the net  is  made of  a large  number or  small,
isolated  graphs or  if  it displays  a  clearly differentiated  Giant
Connected Component  (GCC) that contains  most of the  connected words
-i.e.  words  syntactically active in  some production. The  number of
words  contained  on  such  a  component  or  its  relative  size  are
interesting  statistical   indicators.   Strikingly,  from   the  very
beginning,  child's   syntactic  graphs  display  a   clear  and  very
differentiated GCC. For mathematical  purposes, we will use the matrix
representation of  the connectivity pattern of the  GCC, the so-called
{\em adjacency matrix}.  An element  of such a matrix is $a_{jk}=1$ if
there exists  a link  among the words  $W_j$ and $W_k$  and $a_{jk}=0$
otherwise.  If  the contrary  is not indicated,  -we will  compute the
following measures over the GCC of our graphs.

\noindent
The number  of links  (or {\em degree})  $k_i=k(W_i)$ of a  given word
$W_i  \in {\cal  W}$  gives a  measure  of the  number of  (syntactic)
relations  existing between a  word and  its neighbors.   The simplest
global measure that  can be defined on $\Omega$  is the average degree
$\langle k \rangle$. For the $T$-th corpus, it will be defined as

\begin{equation}
\langle k \rangle_T = {1 \over N_w(T)} \sum_{W_i \in {\cal W}}k(W_i)
\end{equation} 

\noindent where $N_w(T)$ indicates the  number of words present in the
$T$-th corpus.   This number is known to  increase through acquisition
in  a steady  manner.  This  and other  measures are  computed  on the
largest component of the graph.

\noindent
Beyond  the  average  degree,  two  basic  measures  can  be  used  to
characterize   the  graph  structure   of  the   GCC  of   the  $T$-th
corpus. These are  the average path length ($L_T$)  and the clustering
coefficient   ($C_T$).   The   first   is  defined   as  $L_T=   \left
<D_{min}(i,j) \right >$ over all  pairs $W_i, W_j \in {\cal W}$, where
$D_{min}(i,j)$ indicates  the length of the shortest  path between two
nodes. Roughly speaking, a short path  length means that it is easy to
reach a given word $W_i  \in {\cal W}$ starting from another arbitrary
word $W_j \in {\cal W}$. The second is defined as the probability that
two vertices  (e.g.  words) that are  neighbors of a  given vertex are
neighbors of each other. In order to compute the clustering, we define
for each  word $W_i$  a neighborhood $\Gamma_i$.   Each word  $W_j \in
\Gamma_i$  has  been  syntactically  linked  (via  the  above  defined
projection) at least  once with $W_i$ in some  sentence.  The words in
$\Gamma_i$  can  also  be  linked  among  them, and  it  is  what  the
clustering  coefficient evaluates.   The  clustering $C(\Gamma_i)$  of
this set is defined as

\begin{equation}
C(\Gamma_i) = {1 \over k_i(k_i-1)}\sum_j \sum_{k\in\Gamma_i} a_{jk} 
\end{equation} 

\noindent  and  the  average  clustering  of the  GCC  concerning  the
$T$-th corpus   is  simply   $C_T=\langle  C(\Gamma_i)\rangle$.   The
clustering  $C$  provides a  measure  of  the  likelihood of  having
triangles  in the  graph.   Concerning the  average  path length,  for
random graphs with Poissonian structure we have

\begin{equation}
D = 1 + {\log \left [N/z_1 \right ]\over \log \left [z_2 /z_1\right ]}
\end{equation}

\noindent 
being  $z_n$ the  average number  of neighbors  at distance  $n$.  For
Poissonian graphs,  where $z_1=\langle  k \rangle$ and  $z_2=\langle k
\rangle^2$, we have  the following approximation: $D \approx  \log n /
\log  \langle k  \rangle>$.   It is  said  that a  network  is a  {\em
small-world} when $D \approx D_{random}$ (and clearly $D \ll N$).  The
key  difference between  a Poissonian  network and  a real  network is
often $C \gg C_{random}$ \cite{Dorogovtsev}.

\noindent
Another quantity  of interest  is the degree  of affinity  among nodes
with the  same connectivity.   In this way,  the behavior of  hubs is
specially relevant, as well as  they organize the overall structure of
the net. A network is said to  be {\em assortative} if hubs tend to be
connected among them.  At the other side, a network is said to be {\em
dissassortative}   if   hubs   tend   to   avoid   connections   among
them. Language networks  at different scales display a  high degree of
dissassortativeness   \cite{Ferrer04}.    To  quantify   the   degree  of
assortativeness,  we  use  the  so-called  Pearson's  coefficient  for
nets \cite{Assortativitat}:

\begin{equation}
\rho=\frac{c\sum_i j_ik_i-\left(c\sum_i\frac{1}{2}(j_i+k_i)\right)^2}{c\sum_i\frac{1}{2}(j_i^2+k_i^2)-
\left(c\sum_i\frac{1}{2}(j_i+k_i)\right)^2}
\end{equation}

\noindent
where $j_i$ and $k_i$ are the degrees  of the edges at the ends of the
$i$th edge with $i=1,...,m$,  $c=\frac{1}{m}$ and being $m$ the number
of edges. If $\rho<0$ the  net is dissassortative, whereas if $\rho>0$
the net is assortative.

\section{Example}

\noindent
Below we have a fragment of the conversation transcribed in the Corpus
{\em Peter 7}.  We will  detail the analysis that we perform. Firstly,
we  show   the  source  corpus.    We  follow  by   selecting  Peter's
productions.  After that  we select  the structures  and  analyze this
structures and we  tag them. We finish by  computing $\langle {\cal S}
\rangle$ of this fraction of text and by showing the obtained net.

\subsection{The source}

\noindent
*PAT:	hey Pete  that's a nice new telephone  looks like it must do$\\$
$\;\;\;$	everything  it must ring and talk and .$\\$
\%mor:	co|hey n:prop|Pete pro:dem|that v|be $\&$ 3S det|a adj|nice adj|new n|telephone$\\$ 
$\;\;\;$	n|look-PL v|like pro|it v:aux|must v|do pro:indef|everything pro|it v:aux|must $\\$
$\;\;\;$	v|ring conj:coo|and n|talk conj:coo|and . $\\$
\%exp:	Peter has a new toy telephone on table next to him$\\$
\%com:	<bef> untranscribed adult conversation$\\$
*CHI:	xxx telephone go right there .$\\$
\%mor:	unk|xxx n|telephone v|go adv|right adv:loc|there . $\\$
\%act:	<bef> reaches out to lift phone receiver, pointing to place where$\\$
$\;\;\;$	wire should connect receiver and telephone$\\$
*MOT:	the wire .$\\$
\%mor:	det|the n|wire . $\\$
*PAT:	oh <the $\&$ te> [//] the wire's gone ?$\\$
\%mor:	co|oh det|the n|wire v:aux|be $\&$ 3S v|go $\&$ PERF ? $\\$
\%com:	<aft> untranscribed adult conversation$\\$
*CHI:	xxx  need it  my need it  xxx .$\\$
\%mor:	unk|xxx v|need pro|it pro:poss:det|my n|need pro|it unk|xxx .$\\$ 
\%act:	<aft> goes to his room on Mother's suggestion, returns with wire$\\$
*CHI:	xxx .$\\$
\%mor:	unk|xxx . $\\$
*PAT:	uhhuh .$\\$
\%mor:	co|uhhuh . $\\$
*LOI:	why don't you bring your telephone down here  Peter ?$\\$
\%mor:	adv:wh|why v:aux|do neg|not pro|you v|bring pro:poss:det|your n|telephone $\\$
$\;\;\;$	adv|down adv:loc|here n:prop|Peter ? $\\$
*LOI:	why don't you put it on the floor ?$\\$
\%mor:	adv:wh|why v:aux|do neg|not pro|you v|put $\&$ ZERO pro|it prep|on det|the n|floor ?$\\$
\%act:	<aft> Peter puts it on floor <aft> Peter is trying to attack "wire"$\\$
$\;\;\;$	to phone and receiver$\\$
\%com:	<aft> untranscribed adult conversation$\\$
*LOI:	what're you doing ?$\\$
\%mor:	pro:wh|what v|be $\&$ PRES pro|you part|do-PROG ? $\\$
*CHI:	0 .$\\$
\%act:	<aft> Peter goes to hall closet, tries to open it$\\$
*MOT:	what do you need ?$\\$
\%mor:	pro:wh|what v|do pro|you v|need ? $\\$
*CHI:	xxx .$\\$
\%mor:	unk|xxx .$\\$
(...)$\\$
*CHI:	put in there .$\\$
\%mor:	v|put $\&$ ZERO prep|in adv:loc|there . $\\$
\%act:	attaching wire to phone$\\$
*LOI:	ok it's all fixed  oops it was out all fixed  there .$\\$
\%mor:	co|ok pro|it v|be $\&$3S qn|all part|fix-PERF co|oops pro|it v|be $\&$ PAST $\&$ 13S $\\$
$\;\;\;$	adv|out qn|all v|fix-PAST adv:loc|there . $\\$

\subsection{Selected Productions and Analysis}

To  work with  the DGA  Annotator, we  need, firstly,  to  extract the
child's  productions. To  do this,  we  programmed a  routine in  PERL
language able to extract child's  productions. Below there is a simple
pseudocode as a sample:
$\\$

\noindent
\texttt{FILE=PETER}$_k$

\noindent
\texttt{for(i=5; i$<$=LONGFILE; i++)}

\noindent
$\{$

\texttt{if(FILE[i]=~/PETER/)}

$\{$

\texttt{j=j+1;}

\texttt{PETER[j]= "FILE[i]";}

 $\}$

\noindent
$\}$  

$\\$
\texttt{for(i=0; i<=LONGFILE; i++)}

\noindent
$\{$

$\;\;\;\;$  \texttt{@PETER[i]= tr/*PETER:/ /;}

$\;\;\;\;$  \texttt{@PETER[i]=~ tr/./ /;}

$\;\;\;\;$  \texttt{@PETER[i]=~ tr/,/ /;}

$\;\;\;\;$  \texttt{@PETER[i]=~ tr/;/ /;}

$\;\;\;\;$  \texttt{@PETER[i]=~ tr/:/ /;}

$\;\;\;\;$  \texttt{@PETER[i]=~ tr/!/ /;}

$\;\;\;\;$  \texttt{@PETER[i]=~ tr/</ /;}

$\;\;\;\;$  \texttt{@PETER[i]=~ tr/>/ /;}

$\;\;\;\;$  \texttt{@PETER[i]=~ tr/?/ /;}

$\;\;\;\;$  \texttt{@PETER[i]=~ tr/¿/ /;}

$\;\;\;\;$  \texttt{@PETER[i]=~ tr/*/ /;}

\noindent
$\}$
$\\$

\noindent
If we apply the above algorithm  to the sample of text of the example,
we obtain:
$\\$

 xxx telephone go right there

  xxx need it my need it xxx

  xxx

  0

  xxx

  put in there

\subsubsection{XML Format to be read by DGAanotator}

Further we  need to provide the  obtained strings of words  with a XML
format, in order to manage them  with the DGA Annotator. Below we have
an example of the string {\em put in there}.
$\\$

\noindent
$\langle$\texttt{?xml version="1.0" encoding="iso-8859-1"}$\rangle$

\noindent
$\langle$\texttt{!DOCTYPE DGAdoc SYSTEM "dga.dtd"}$\rangle$

\noindent
$\langle$\texttt{DGAdoc}$\rangle$

\noindent
$\langle$\texttt{s}$\rangle$

$\langle$\texttt{tok}$\rangle$

$\;\;\;\;\;\;\;\langle$\texttt{orth}$\rangle$\texttt{put}$\langle$\texttt{/orth}$\rangle$

$\;\;\;\;\;\;\;\langle$\texttt{ordno}$\rangle$\texttt{1}$\langle$\texttt{/ordno}$\rangle$

$\langle$\texttt{/tok}$\rangle$

$\langle$\texttt{tok}$\rangle$

$\;\;\;\;\;\;\;\langle$\texttt{orth}$\rangle$\texttt{in}$\langle$\texttt{/orth}$\rangle$

$\;\;\;\;\;\;\;\langle$\texttt{ordno}$\rangle$\texttt{2}$\langle$\texttt{/ordno}$\rangle$

$\langle$\texttt{/tok}$\rangle$

$\langle$\texttt{tok}$\rangle$

$\;\;\;\;\;\;\;\langle$\texttt{orth}$\rangle$\texttt{there}$\langle$\texttt{/orth}$\rangle$

$\;\;\;\;\;\;\;\langle$\texttt{ordno}$\rangle$\texttt{3}$\langle$\texttt{/ordno}$\rangle$

$\langle$\texttt{/tok}$\rangle$

\noindent
$\langle$\texttt{/s}$\rangle$

\noindent
$\langle$\texttt{/DGAdoc}$\rangle$
$\\$

\begin{figure}
  \includegraphics[width=7.5 cm]{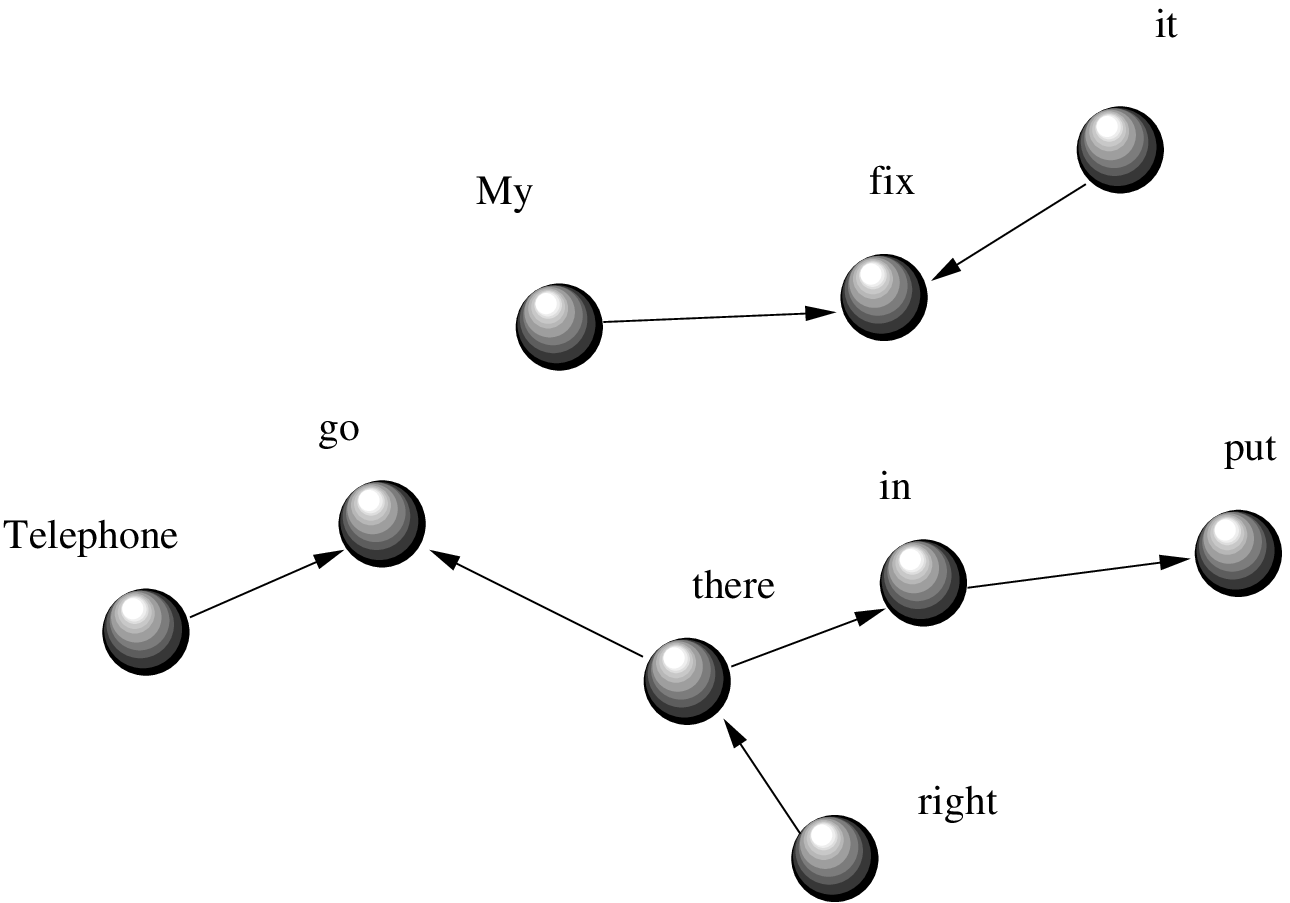}
  \caption{Graph of the sample.}
  \label{petitgraf}
\end{figure}

\subsubsection{Selection of {\em valid} strings, annotation and computation of ${\cal S}$}

\noindent
We reject $xxx$ and $0$ as  lexical items and proceed to annotate with
the DGA anotator.

Once the corpus is annotated  (with the criteria developed through the
paper!) we generate the set of words. This can be done by sampling the
XML file  once annotated by  using a routine  close to the  ones shown
above  (PERL or  Python are  the ideal  languages).  To  compute graph
parameters and more mathematical artifacts, it is a good choice to use
a {\em stronger} language, such as C or C++.

\begin{eqnarray} 
{\cal W}=\left\{ {\rm telephone,\;go;\; right,\;there,\;need,\;it,\;my},\right. \nonumber \\
\left.{\rm  put,\;in} \right\}
\end{eqnarray}

\noindent 
And the analysis is, roughly speaking:

\begin{equation}
s_1= {\rm[ telephone [go[right\;there]_{PP}]_{VP}]_{TP}}\;\;|s_1|=4
\end{equation}

\begin{equation} 
s_2={\rm [need\; it]_{VP}}\;\;|s_2|=2
\end{equation}

\begin{equation} 
s_3={\rm [my [need \;it]_{VP}]_{TP}}\;\;s_3=3
\end{equation}

\begin{equation} 
s_4={\rm [put [in \;there]_{PP}]_{VP}}\;\;s_4=3
\end{equation}

\noindent
Thus, we can compute $\langle {\cal S}\rangle$:

\begin{equation}
\langle {\cal S}\rangle=\frac{4+2+3+3}{4}=3
\end{equation}

\noindent
and, following the criteria developed above, we can define ${\cal E}$

\begin{eqnarray}
{\cal E}=\left\{ {\rm telephone \rightarrow go,\; right \rightarrow here,\;here\rightarrow go;}  \right. \nonumber\\ 
\left.{\rm it \rightarrow need,\; my \rightarrow need, \;there \rightarrow in,\; in \rightarrow put} \right\}  
\end{eqnarray}

\noindent
We have built the graph.

\section{Acknowledgments}
My  indebtedness for  Carlos Rodr\'iguez-Caso  for his  ideas  and his
patient advising in  teaching PERL. I also want  to acknowledge Harold
Fellermann  for  his  help  in programming  python  routines,  Andreea
Munteanu for the careful reading of the manuscript and Barbara Soriano
for  its   useful  comments  on  language   acquisition  process.   My
indebtness, also,  to Ricard V. Sol\'e  and Sergi Valverde  as well as
this  work is  a  part of  a bigger  work  where their  ideas have  been
illuminating. Finally,  I'd like to acknowledge Ramon  Ferrer i Cancho
for his contributions at the beginning of the work.

\end{document}